\documentclass{svproc}

\usepackage{url}
\usepackage{hyperref}
\usepackage{graphicx}
\usepackage{placeins}
\usepackage{array}
\usepackage[table]{xcolor}

\begin{document}
\mainmatter              
\title{Question: How do Large Language Models perform on the Question Answering tasks? Answer:}

\titlerunning{How do LLMs perform on the QA task?}  
\author{Kevin Fischer\thanks{Equal contribution.}  \and Darren Fürst\footnotemark[1] \and Sebastian Steindl  \and
Jakob Lindner \and Ulrich Schäfer}

\authorrunning{Kevin Fischer, Darren Fürst et al.}
\tocauthor{Kevin Fischer, Darren Fürst, Sebastian Steindl, Jakob Lindner, and Ulrich Schäfer}

\institute{
\email{\{k.fischer2, d.fuerst, s.steindl, j.lindner, u.schaefer\}@oth-aw.de}
\\
 Ostbayerische Technische Hochschule Amberg-Weiden\\
92224 Amberg, Germany
}

\maketitle              

\begin{abstract}
Large Language Models (LLMs) have been showing promising results for various NLP-tasks without the explicit need to be trained for these tasks by using few-shot or zero-shot prompting techniques. A common NLP-task is question-answering (QA). In this study, we propose a comprehensive performance comparison between smaller fine-tuned models and out-of-the-box instruction-following LLMs on the Stanford Question Answering Dataset 2.0 (SQuAD2), specifically when using a single-inference prompting technique. 
Since the dataset contains unanswerable questions, previous work used a double inference method.
We propose a prompting style which aims to elicit the same ability without the need for double inference, saving compute time and resources.
Furthermore, we investigate their generalization capabilities by comparing their performance on similar but different QA datasets, without fine-tuning neither model, emulating real-world uses where the context and questions asked may differ from the original training distribution, for example swapping Wikipedia for news articles.

Our results show that smaller, fine-tuned models outperform current State-Of-The-Art (SOTA) LLMs on the fine-tuned task, but recent SOTA models are able to close this gap on the out-of-distribution test and even outperform the fine-tuned models on 3 of the 5 tested QA datasets.

\end{abstract}
\section{Introduction}
Natural Language Processing (NLP) as a subfield of Artificial Intelligence deals with processing human natural language through algorithms. 
The widespread adoption of LLMs has greatly influenced the NLP community, both in academia and industry.
These LLMs show strong natural language generation and understanding capabilities, thanks to their vast number of parameters coupled with extensive pretraining on very large amounts of tokens.
Additonally, LLMs have prompt-following capabilities that is achieved with instruction-tuning \cite{zhangInstructionTuningLarge2023}.

A user can therefore describe a task in natural language and give context, and the LLM will try to output a fitting answer. This can work without fine-tuning the model for the specific task at hand, which could eliminate the need for specialized models, bearing the potential to greatly reduce effort for data curation and model training.

LLMs are therefore being investigated on various tasks, such as summarization \cite{zhang2024benchmarking}, named entity recognition \cite{wang2023gpt}, but also as judges during evaluation \cite{zheng2023LLMJudge} and even time-series forecasting \cite{jin2023time}, to name a few.

This has led to research into prompting techniques \cite{kojima_large_2023,qin_is_2023}, where a natural language prompt describes the context, task and gives examples.

One common NLP task is question-answering (QA), which is of great interest for both academia and applications in industry.
In this task, the model has to derive the answer to a question based on a given context.
Fine-tuned transformer-based models perform well on this and established the SOTA \cite{Rawat2022Comparative}. Yet, this fine-tuning needs resources and is naturally limited in generalization to different domains or styles of question.

We therefore investigate whether prompt-following LLMs can be used out-of-the-box without any fine-tuning to solve the QA task and how well they perform compared to smaller models that went through task-specific fine-tuning.
To this end, we evaluate models fine-tuned for the specific task on the SQuAD2 dataset and on external QA datasets to understand their level of out-of-distribution generalization.

The SQuAD2 dataset contains questions both answerable and unanswerable given the context.
The latter are of special interest from the perspective of hallucinations.
We derive a special prompt (cf. Appendix) to treat both types of questions in one single inference, reducing the computing time and cost.
We compare the LLMs  LLaMA2 7B \cite{touvronLLaMAOpenFoundation2023}, LLaMA3.1 8B, LLaMA3.1 70B, LLaMA3.2 1B, LLaMA3.2 3B \cite{llama3} and GPT-4 Turbo out-of-the-box, against the three smaller, fine-tuned models Flan-T5 \cite{chung2022scaling}, DistilBERT \cite{sanh2020distilbert} and RoBERTa \cite{liu2019roberta}.

In conclusion, our main contributions are the empiric comparison between multiple LLMs, both open-source and proprietary, with different fine-tuned models on the SQuAD2 dataset and external datasets, thus evaluating generalization capabilities. Furthermore, we propose a single-inference prompt, that allows to treat both answerable and unanswerable questions with one forward pass instead of two, halving the inference steps.
Lastly, we extend upon the classic QA evaluation metrics by a new approach, taking into account how far from the correct answer the models are based on the Levenshtein Distance and analyze the performance for the different interrogative pronouns.

Our results indicate, that sufficiently large LLMs can achieve competitive results on QA tasks out-of-the-box and can thus be considered as an alternative to fine-tuned specialized models. While they require less resources to prepare (i.e. no fine-tuning), they incur higher inference-costs due to their size.

\section{Background and Related Work}
While early question-answering systems (QAS) like Baseball \cite{Baseball} or LUNAR \cite{LUNAR} were able to answer questions in a very specific domain, the SQuAD \cite{rajpurkar2016squad} dataset contains more than 100,000 questions based on paragraphs from Wikipedia articles and has become a widely used benchmark for QAS. Its second version \cite{rajpurkar2018know} introduces unanswerable questions, additionally. These are related to the given context, but cannot be answered from it.

Different models and architectures have been evaluated on this benchmark. Pearce et al.~\cite{pearce2021comparative} compared different versions of BERT on SQuAD2. RoBERTa achieved the best F1-score with 68.2. Özkurt~\cite{ozkurt2024comparative} did a similar comparison, but with models that were fine-tuned for three epochs. Here, ALBERT achieves an F1-score of 89.9, which is 7 points better than RoBERTa in this study.

Since its publication, ChatGPT has been increasingly studied in the context of QA as well, including the SQuAD and SQuAD2 dataset.
Kocon et al.~\cite{kocon_chatgpt_2023} conducted an extensive study on the performance of ChatGPT in different NLP tasks, including perfomance on QA-tasks. Part of it is a comparison with the SOTA, DeBERTaV3, on SQuAD, where the SOTA (F1: 90.75) achieved a much better result than ChatGPT (69.21).
An analysis by Bahak et al.~\cite{bahak2023-iranisches-paper} compared ChatGPT3.5 \cite{brown2020language} with other pretrained LLMs and task-specific models for QA on different datasets, including SQuAD2. ChatGPT fell behind in the comparison, but its results could be improved by providing more context and prompt engineering, especially by the introduction of two-step queries.
Omar et al.~\cite{omar2023chatgpt} carried out a comparison of ChatGPT with the traditional QAS KGQAN \cite{omar2023universal} and EDGQA \cite{edgqa} for knowledge graphs. ChatGPT showed convincing results in explainability, question understanding, and robustness, but had issues in correctness, incorporating recent information, and generality, where the traditional QASs performed better.

Tan et al.~\cite{tan2023chatgpt} conducted an extensive comparison of the LLMs ChatGPT, GPT-4, GPT-3.5, and FLAN-T5 with traditional knowledge-based QA models on six English and two multilingual benchmark datasets, not including SQuAD. FLAN-T5 achieved similar results as GPT, but the fine-tuned SOTA was still better in most tests.
\section{Method and Experiments}
To investigate our main research question, we compared variants of DistilBERT, Flan-T5, the LLaMA-model family and GPT-4 Turbo on the SQuAD2 dataset. We measure the performance of each model on common QA metrics such as F1-score and Exact Match. Additionally, we perform an extended evaluation, where we investigate how far off the model prediction was from the ground truth utilizing the Levenshtein Distance, secondly compare the generalization to other QA datasets and lastly break down the results by interrogative pronoun.
A special complexity of the SQuAD2 dataset are adversarially crafted questions, which are unanswerable from the given context to test the model's reading comprehension \cite{rajpurkar_know_2018}. These seem to be answerable, but the context does not actually include the answer.

\subsection{Models}
In this study, the LLMs GPT-4 Turbo (version gpt-4-0125-preview, in the following referred to as GPT4) and various open-source LLaMA-models are compared to the fine-tuned Language Models Flan-T5 \cite{chung2022scaling}, DistilBERT \cite{sanh2020distilbert} and RoBERTa \cite{liu2019roberta}, in their ability of working as a QAS. For the sake of readability, GPT4 and the LLaMA-models will be collectively referred to as LLMs, and the three fine-tuned models as FT-LMs. 
The LLMs were used with a temperature of 0.1 and a repetition penalty of 1.1 to encourage a concise response close to the context. 

In comparison to the LLMs, which were instructed to perform as QAS in the prompt but are otherwise used out-of-the-box, all the FT-LMs were fine-tuned on the SQuAD2 training dataset. Flan-T5\footnote{Flan-T5 weights from \href{https://huggingface.co/sjrhuschlee/flan-t5-base-squad2}{Huggingface}} has 223 million parameters and was trained for 4 epochs and with a batch size of 16, using Adam \cite{kingma2017adam} with a learning rate of $2*10^{-5}$ \cite{chung2022scaling}. RoBERTa\footnote{RoBERTa weights from \href{https://huggingface.co/deepset/RoBERTa-base-squad2}{Huggingface}} was fine-tuned for 2 epochs on 4 Tesla v100 with a batch size of 96 and a learning rate of $3*10^{-5}$ and 124 million parameters \cite{liu2019roberta}. The DistilBERT\footnote{DistilBERT weights from \href{https://huggingface.co/twmkn9/DistilBERT-base-uncased-squad2}{Huggingface}} version used here was trained for 3 epochs, with a batch size of 8 and a learning rate of $3*10^{-5}$. With its 67 million parameters \cite{sanh2020distilbert}, it's the smallest model used for this comparison.
While the number of GPT4's parameters is not publicly available, the used open-source models range from  1-70 billion parameters.

\subsection{Prompting Strategy}
To test how well the LLMs can perform on this task with few-shot learning, a prompt that uses a persona strategy \cite{white2023prompt} was developed to instruct the model on the reading task by telling it that it is a reading comprehension assistant.
Previous work has shown that using a two-step inference technique allows for better filtering of unanswerable questions \cite{bahak2023-iranisches-paper}.
This works by first asking the model for a binary response to see if the question seems answerable or not and only instructing it to answer if it deems the question to be answerable \cite{bahak2023-iranisches-paper}.
Our approach aims to achieve similar results with only a single inference step to cut down on time and (resource) costs.
To this end, we provide the model with an algorithm to base its answer on and telling it to "think step-by-step", which has been shown to elicit Chain-of-Thought (CoT) reasoning in LLMs \cite{kojima_large_2023} and can thus improve performance on the QA task.

\subsection{Unanswerable Questions}
The benchmark dataset contains questions, to which no answer can be derived from the given context. They are thus unanswerable and increase the difficulty of the task. 
For prompt-following LLMs, this is of special interest regarding hallucinations, where the model outputs a factually wrong but seemingly right answer. Additionally, since these models benefit of pretraining on large amounts of data, they have encoded certain world-knowledge in their parameters, which in general allows them to answer factual questions to some extent. Therefore, it is possible that the LLM might be able to give an answer based on its pretraining data to a question that is unanswerable given the context. 

To handle these questions, we prompted the LLM to answer with only the string \texttt{"unanswerable"} for unanswerable questions. In the prompt, the LLMs were also instructed to answer as concisely as possible, as this is very important when wanting a specific answer to a question without extra information, which is the expected format for the evaluation. 
Even though the LLMs had almost the same prompt, we found the LLaMA models to follow the instruction less reliably than GPT4. For example, they tried to explain why the question was not answerable. 

The FT-LMs' answers were counted as unanswerable if their predicted probability for this was above a certain threshold, which is a standard practice when using these models. The official evaluation script\footnote{\href{https://github.com/huggingface/evaluate/blob/main/metrics/squad_v2/compute_score.py}{SQuAD2 evaluation script}} provides separate thresholds for the No-Answer-Probability that either work best for EM or F1. The threshold used for this evaluation was chosen as the overall mean between the best-achieved F1 threshold and the best-achieved EM threshold.
Table \ref{tab:results} shows the results for the unanswerable questions separately. While the FT-LMs were able to achieve good results, the EM score for the LLMs is remarkably lower in most cases.

\section{Evaluation}
To get a good understanding of the capabilities of LLMs compared to FT-LMs on the QA task, we evaluate the models with the Exact Match (EM) and F1 score metrics on the official SQuAD2 validation split, which was not used during the fine-tuning of FT-LMs.
Furthermore, we detail the evaluation regarding the type of question and how the score would change if a certain amount of character changes to the answer were allowed. This aims to identify close-misses, that would be semantically correct, but are marked as incorrect due to the exact string comparison.

The EM score is a strict evaluation metric that measures whether a model's response exactly matches one answer in the given ground truths. It is commonly used because it provides a clear, binary indicator of a model's accuracy in reproducing precise answers, making it straightforward to assess and compare the performance of different models. 
The F1 score is a statistical measure used to evaluate the accuracy of a model by calculating the harmonic mean of precision and recall. The F1 is defined by \[ F1 = \frac{2* Precision * Recall}{Precision + Recall} \] where Precision is the amount of correct words in the model's prediction and Recall is the amount of words in the ground truth that are also in the model's prediction.

To get a more detailed understanding of the answer quality, we also calculate the Levenshtein Distance (LD) between each model's prediction and the best matching answer. LD is a metric for measuring the difference between two strings by counting the minimum number of single-character edits (insertions, deletions, or substitutions) required to change one sequence into the other. 

Since all of these metrics are string based, the official evaluation procedure applies some preprocessing to the model responses to normalize them before calculating the scores. This includes, e.g., removing stopwords such as ``the'' and punctuation, ensuring that answers are not flagged as false negatives due to punctuation which does not influence the semantic correctness. 

Due to their nature, LLMs are especially prone to produce semantically correct responses, that would fail a string-based comparison, because they tend to include additional tokens by prepending introductory phrases such as ``Sure, your answer is'' or closing statements like ``Is there anything else I can do for you?''. 
This is an artifact of their training process, where more verbose answers with these 
phrases are seen as preferable towards a \textit{friendly and helpful} chatbot. 
We therefore apply a very light post-processing to these answers, detailed in section~\ref{sec:postProcessLLM}.

\subsection{Evaluation on the Benchmark}
\begin{table}[!ht]
\caption{\label{tab:results}
EM and F1 score of each model on all questions of the SQuAD2's official validation dataset. The EM score on exclusively unanswerable questions is shown separately in the column ``noAns EM''. All LLMs are the instruction-tuned version.}
\centering
\rowcolors{2}{gray!15}{white}

\begin{tabular}{>{\raggedright\arraybackslash}p{3.7cm}>{\centering\arraybackslash}p{2cm}>{\centering\arraybackslash}p{2cm}>{\centering\arraybackslash}p{2cm}}
\rowcolor{gray!30} 
\textbf{Model} & \textbf{EM} & \textbf{F1} & \textbf{noAns EM}\\
\hline
Flan-T5              & 64.55 & 66.02 & 59.7\\
DistilBERT           & 67.89 & 70.17 & 72.2\\
RoBERTa              & 79.97 & 82.43 & 83.48\\
GPT4-Turbo           & 46.48 & 59.56 & 49.12\\
LLaMA-2-7B           & 43.85 & 50.09 & 40.84\\
LLaMA-3.1-8B & 41.01 & 47.22 & 13.71\\
LLaMA-3.1-70B        & 57.13 & 73.68 & 81.56\\
LLaMA-3.2 1B  & 20.97 & 28.28 & 1.30\\
LLaMA-3.2 3B & 45.68 & 52.72 & 31.34\\
\end{tabular}
\end{table}

Our main evaluation is performed on the SQuAD2 dataset, which is a widely used QA benchmark. The results on this benchmark are shown in Table~\ref{tab:results}, where the EM for exclusively unanswerable questions is additionally shown separately. While each LLM's answer is represented as a single string, the FT-LM's answers also include a No-Answer-Probability-Score (NAP) as well as the start and end indexes of the tokens that represent the answer. The NAP is used to decide, if the model thinks the question is answerable or not.

To allow the evaluation and its metrics to focus on the key content of each answer, both prediction and ground-truth are normalized. To this end, the answer is transformed to lower case, followed by the removal of all punctuation marks. In the resulting strings, the articles ``a", ``an" and ``the" were removed. Finally, a white-space-fix reduced several consecutive white spaces to a single one.
This is the standard procedure for evaluating on the SQuAD2 dataset, as defined by the official evaluation script.
Since the dataset contains multiple ground-truths, we report the best metric over all ground-truths.

The evaluation in \cite{llama3} utilizes previous questions and answers on a document as additional context within their prompt. Even though they achieve higher scores in this manner, it would be unfair in a comparison with FT-LMs. We therefore do not utilize this prompting strategy, but instead compare all LLMs without the previous questions as context.

\subsection{Extended Evaluation}
To get a deeper insight into the models' abilities in QA, each model's F1 and EM-scores were compared by the type of questions contained in SQuAD2. The questions were sorted into the interrogative pronouns ("what", "how", "when", "where", "who", "which", "why" and "other") by first appearance. If none of them were found, the question was categorized as "other". 

\begin{figure}[!ht]
    \centering
    \includegraphics[width=\linewidth]{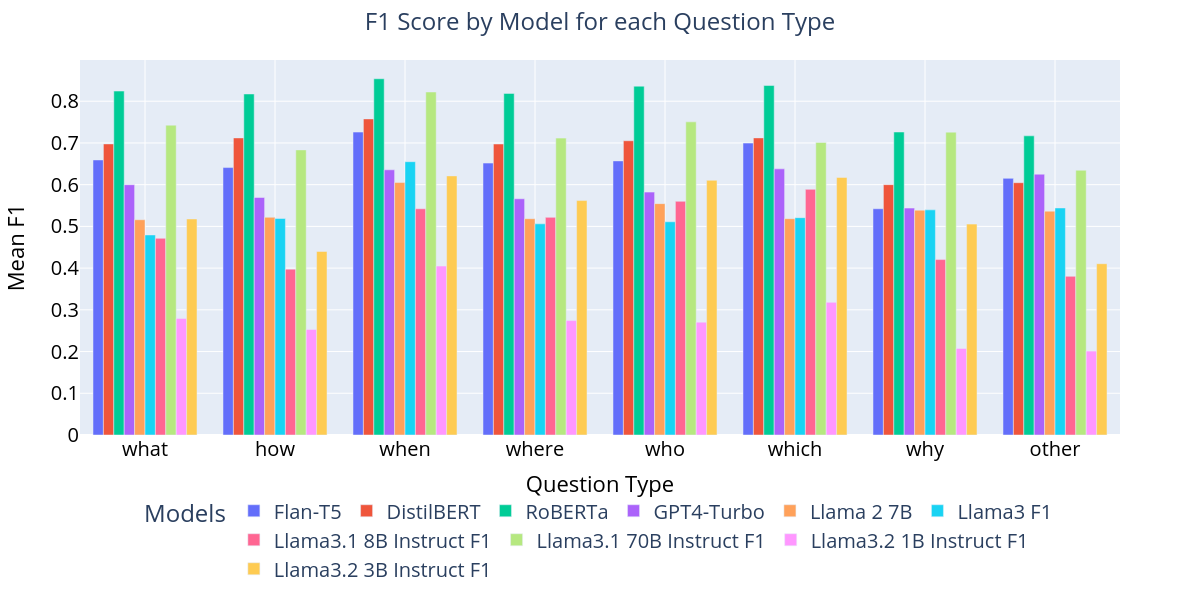}
    \caption{The mean F1 score of the models discerned by interrogative pronouns.}
    \label{fig:question-types}
\end{figure}

The F1-scores shown in Fig. \ref{fig:question-types} per interrogative pronoun and model show that while RoBERTa was able to achieve the overall best score for all possible question types, LLama3.1 70B Instruct was the second-best model for all pronouns except "how" and "which", even rivaling RoBERTa for the pronoun "why". 
Furthermore, we can see that all models perform relatively constant over all interrogative pronouns, so no remarkable bias exists. It is however interesting to see that all models perform relatively worse on ``why'' questions. 
This might hint to limited reasoning capabilities for causal questions. and is in line with previous studies \cite{bahak2023-iranisches-paper}.

These results indicate that the recent advancements in LLMs are starting to be able to rival fine-tuned smaller models, if prompted correctly.
Regarding the type of question, we can see for all models that they perform relatively constant over all interrogative pronouns, so no remarkable bias exists. It is however interesting to see that all models perform relatively worse on ``why'' questions. 
This might hint to limited reasoning capabilities for causal questions. 

\begin{figure}[!ht]
    \centering
    \includegraphics[width=\linewidth]{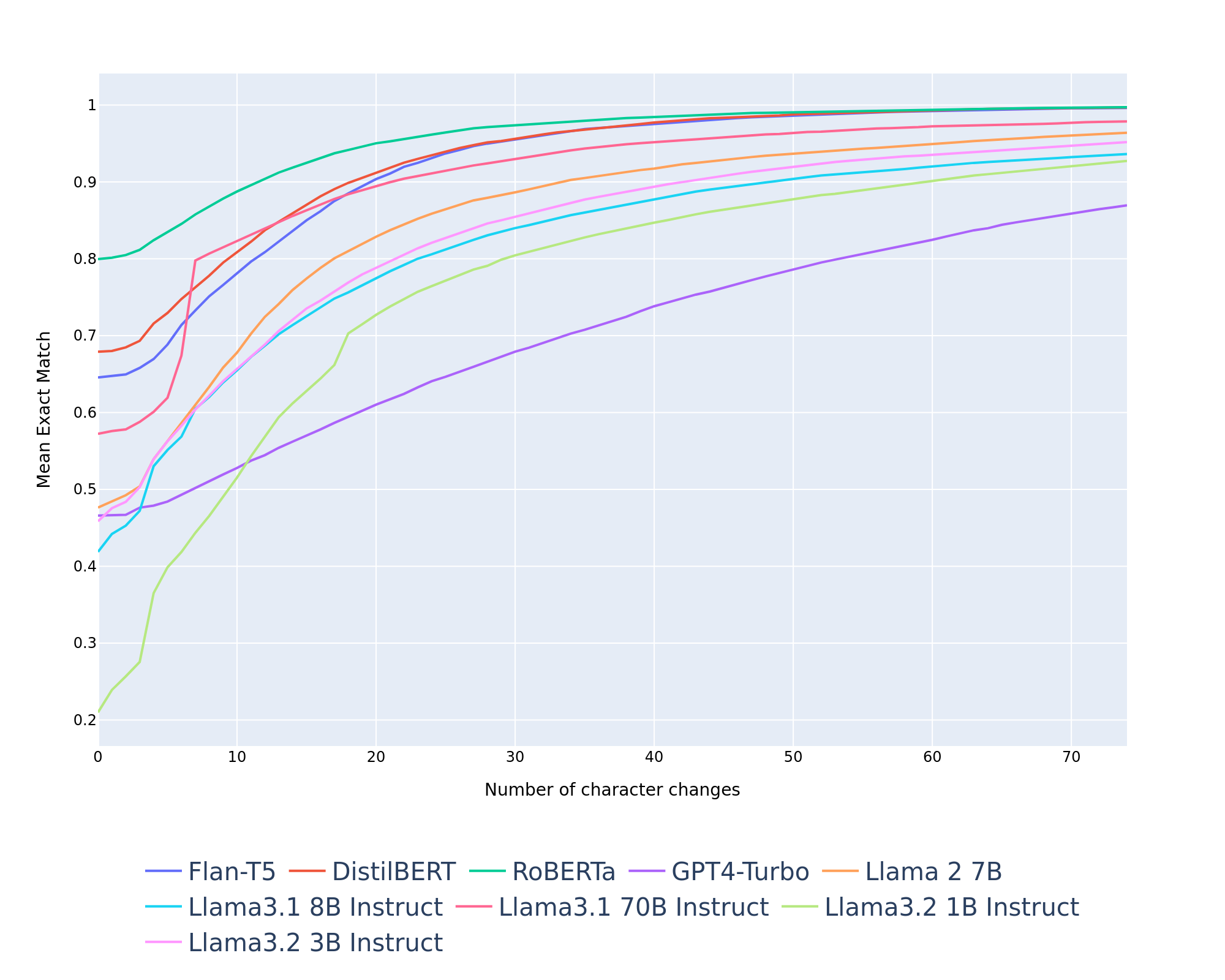}
    \caption{Mean EM score if $n$ character changes (cf. Levenshtein Distance) to the answer were allowed.}
    \label{fig:lev-changes}
\end{figure}

Furthermore, the improvement of the EM score was investigated if $n$ character changes were allowed. For this purpose, the Levenshtein distances between the ground truth and the model answers were calculated (cf. Fig.~\ref{fig:lev-changes}).
This distance is meant to give a sense of how far away the models were from the right answer.

This metric shows how close the models were to the correct answer semantically and can give an intuition, especially for LLMs, whether the model response deviates only slightly from the ground-truth, e.g., in used articles or number.
Since the evaluation compares character-wise, small syntactic changes might give a false negative for a semantically correct answer.
While a large $n$ will allow changing a completely wrong answer to the right one, evaluating all models in the same scenario will still give a fair comparison, and our focus lies specifically on smaller $n$.
The results show that allowing the models 2-3 character changes didn't lead to a meaningful improvement. This is because of the removal of articles in the answer normalization. Allowing up to 20 character changes made a big difference for all models. 
An interesting case is the LLaMA-3.1-70B-Instruct model, which has a sharp improvement in EM score at 8-10 character changes. This is a clear indicator that more elaborate post-processing could have improved its score.

\subsection{Post-Processing for LLM predictions}\label{sec:postProcessLLM}
Because of the training procedure of the LLMs, they tend to answer verbosely. 
This can to some extent be prevented by specifying within the prompt, that the response should only include the answer to the question and nothing else (cf. the full prompts in the appendix).

However, we still found the LLMs, especially of the LLaMA-family, to include unnecessary tokens in their response. These ranged from courtesy text to pseudo-special tokens such as ``[\textbackslash ANSWER]'', which is analog to LLaMA-specific special token ``[\textbackslash INST]'' which marks the end of the instructions. 

We therefore apply light post-processing to the LLM answers based on regular expressions and a qualitative investigation of the different types of extraneous tokens. 
This has greatly improved all calculated scores for the open-source LLMs.
We found GPT4 to adhere to the prompt close enough and not produce extraneous tokens as to need post-processing. This can also be seen in Fig.~\ref{fig:lev-changes}.

As the improvement of the EM score for a small number of character changes $n < 10$ in Fig.~\ref{fig:lev-changes} clearly indicates, a more intensive post-processing could have further improved the performance.
However, to keep the comparison to FT-LLMs fair, we did not perform detailed model-specific post-processing.
\section{Out of Distribution Comparisons}
Previously, it could be seen that LLMs are in most cases unable to perform on the level of FT-Models on the SQuAD2 dataset. However, one of the advantages of LLMs is their ability to adapt to new tasks in a zero-shot or few-shot setting. In real-world applications where tasks may differ, it can be difficult to fine-tune models in the first place, or have them generalize well to the different tasks at hand. 
Since the LLMs were not adapted to the SQuAD2 dataset in the first place, one can expect that their performance would be similar on other QA datasets in a similar format, meaning a better out-of-distribution (OOD) generalization, with the SQuAD2 dataset being the in-distribution.

\begin{figure}[!ht]
    \centering
    \includegraphics[width=\linewidth]{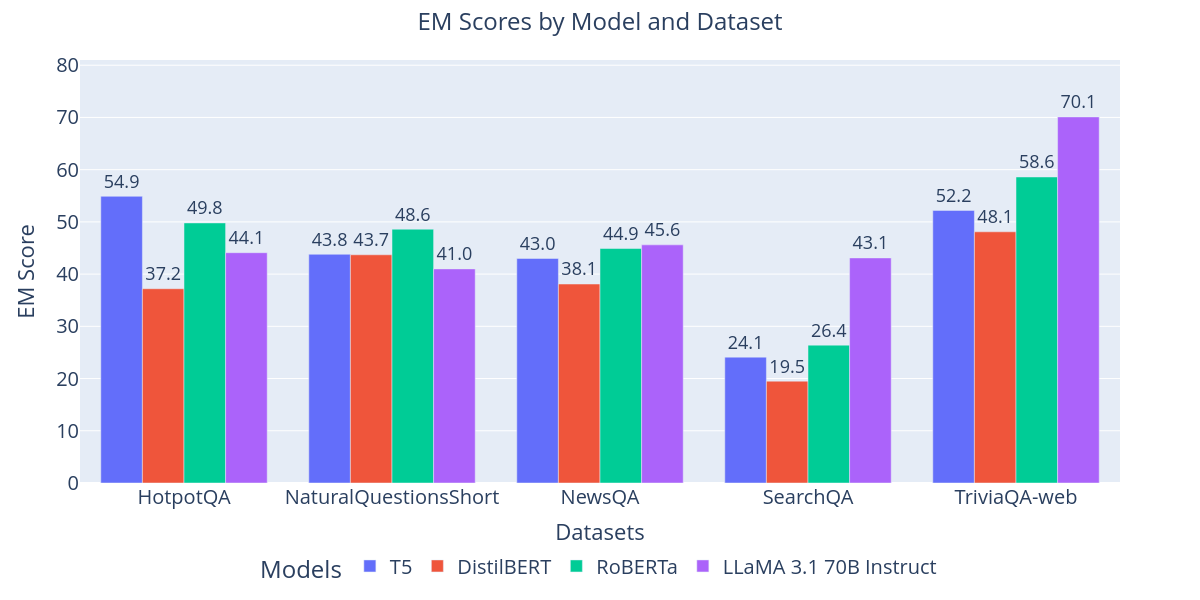}
    \caption{EM-Scores of each model on the OOD-Datasets.}
    \label{fig:em-ood}
\end{figure}

\begin{figure}[!ht]
    \centering
    \includegraphics[width=\linewidth]{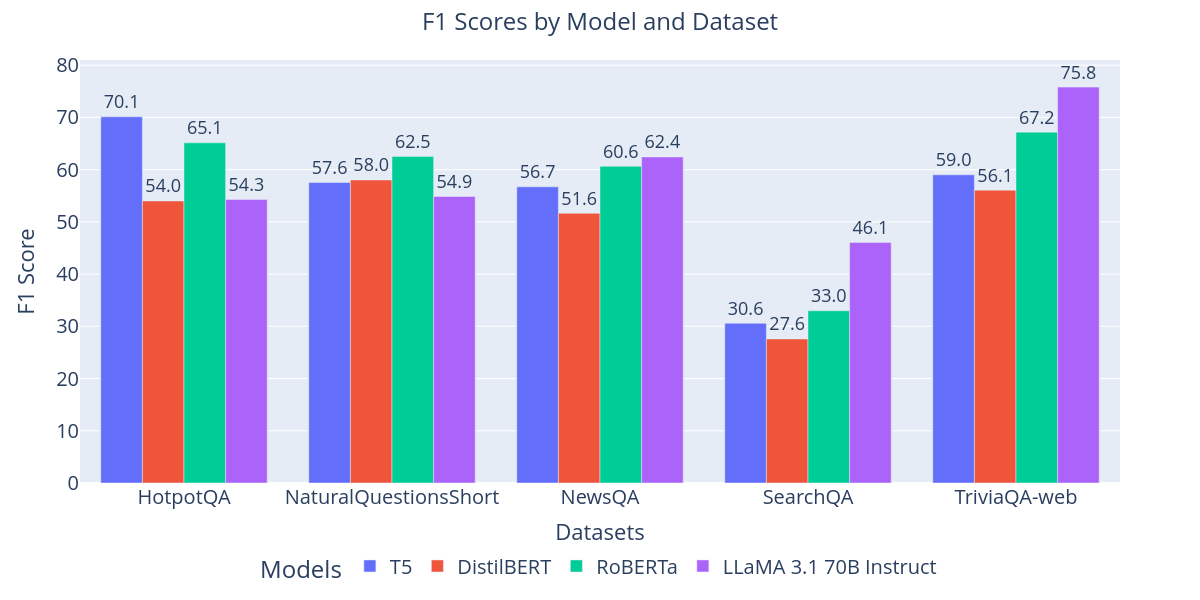}
    \caption{F1-Scores of each model on the OOD-Datasets.}
    \label{fig:f1-ood}
\end{figure}

To evaluate this hypothesis, we test the best performing LLM, LLAMA 3.1 70B Instruct, and the FT-Models on 1000 questions each of five further QA datasets. The results as seen in Fig.~\ref{fig:em-ood} and Fig.~\ref{fig:f1-ood} show that the FT-Models also generalize to some extent to these datasets. Similar to before, RoBERTa is the best FT-LM in four out of the five tested OOD datasets. However, the LLM outperforms all FT-LMs on three of the five datasets.

\section{Results and Discussion}
The results in Tab.~\ref{tab:results} show that RoBERTa was able to achieve the best results in both the F1 and the EM score.
RoBERTa achieved more than 10 percentage points improvement over the other FT-LMs. DistilBERT was slightly better than Flan-T5 in both metrics. The LLMs were in general not able to perform as well as even the worst FT-LM on the EM-scores, however, LLaMA-3.1-70B-Instruction achieved the second-best F1-score. GPT4 is almost 3 percentage points better than LLaMA-2 in the EM metric and 9 percentage points better in the F1 score. The same pattern can be observed in the EM for questions that were not answerable.
However, with the newer generation of LLaMA-models, we can see that LLaMA-3.1-70B achieves the best scores of all LLMs and the significantly smaller LLaMA-3.2 3B comes close to the performance of GPT4, albeit the latter performs remarkably better for unanswerable questions.
Additionally, LLaMA-3.1-70B achieves results comparable to RoBERTa in the F1 score and the EM score for unanswerable questions.
Moreover, the LLaMA-3.2-1B has a very low EM on the unanswerable questions.

\begin{table}
\centering
\caption{\label{tab:levenshteinDist}
The mean $\bar{x}$ and median $\widetilde{x}$ Levenshtein Distances for all questions ($\bar{x}_{all}$,  $\widetilde{x}_{all}$) and the impossible questions ($\bar{x}_{imp}$).}
\rowcolors{2}{gray!15}{white}

\begin{tabular}{>{\raggedright\arraybackslash}p{2.2cm}>{\centering\arraybackslash}p{2cm}>{\centering\arraybackslash}p{2cm}>{\centering\arraybackslash}p{2cm}}
\rowcolor{gray!30} 
\textbf{Model} & $\bar{x}_{all}$ & $\widetilde{x}_{all}$ & $\bar{x}_{imp}$\\
\hline
Flan-T5 & 6.11 & 0 & 6.85 \\
DistilBERT & 5.48 & 0 & 4.71 \\
RoBERTa & 3.36 & 0 & 2.74 \\
GPT4-Turbo & 28.56 & 7.00 & 32.4 \\
LLaMA-2-7B & 12.96 & 3.00 & 13.42 \\
LLaMA3.1 8B   & 19.86 & 4.00 & 30.34 \\
LLaMA3.1 70B   & 8.41 & 0 & 6.77 \\
LLaMA3.2 1B   & 21.57 & 10.00 & 28.18 \\
LLaMA3.2 3B   & 15.94 & 3.00 & 21.98 \\
\end{tabular}

\end{table}

Tab. \ref{tab:levenshteinDist} shows statistics for the Levenshtein Distance. This shows that the LLMs answers are \emph{wronger} and / or longer than those of the FT-LMs, since the distances are remarkably higher.
This can be explained by the LLMs answering in free text, while the FT-LMs answer index-based, and can thus not generate extraneous tokens by design.
The mean distance between the correct answer and the prediction from the model is double that for LLaMA-2 than for Flan-T5. For GPT4, it's again twice the distance compared to LLaMA-2. As before, RoBERTa had the best results, with a mean distance of 3.36. While LLaMA-3.1-70B still has a higher distance than all FT-LMs, it is by far the closest of all LLMs with a distance of 8.41.

To exclude outliers, the median was calculated. This shows that all the medians for the FT-LMs are zero, while LLaMA-3.1-70B is the only LLM achieving this. Furthermore, this metric shows that even if a LLaMA-family model had worse metrics in EM and F1 than GPT4, its wrong answers are on average closer to the ground truth than those of GPT4.
This effect also shows in \ref{fig:lev-changes}, where LLaMA-2 quickly takes over GPT4 once a few character changes are allowed.
From this, we can conclude that GPT4 stayed closer to the prompt and performs better in a strict evaluation, while many of the LLaMA models perform better in a fuzzy evaluation, because their wrong answers are closer to the right ones.

\section{Limitations}
This work evaluates a specific NLP task combining reading comprehension and QA. To improve the understanding of the models' performances, we include an OOD evaluation. Nevertheless, our results should not be expected to be transferable to other NLP tasks, where the balance between either FT-LM and LLM or within the specific models might be completely different. 
While we use carefully crafted prompts and empirically evaluate its effectiveness on these models, there may well exist a prompt that provides better results.

\section{Conclusion and Future Work}
In this paper, we investigated how LLMs perform on QA-tasks compared to models specifically fine-tuned for this purpose. To conduct this study, we performed an empirical comparison on the SQuAD2 dataset, as well as on out-of-distribution testing on five further QA-datasets.

Our results indicate that most LLMs are less effective than FT-LMs in answering SQuAD2 questions, consistent with findings from prior research. However, one notable exception is LLaMA-3.1-70B Instruct, which achieved the second-highest F1 score, only trailing behind RoBERTa. This result suggests that recent advancements in LLM architecture and prompting techniques have enabled some LLMs to outperform smaller FT-LMs when provided with an optimized prompt.
However, one has to keep in mind that this comes at the cost of a large amount of model parameters and therefore resources needed for inference.

We also observed that LLMs struggle particularly with unanswerable questions, likely due to their known tendency to hallucinate. Additionally, all models, regardless of training type, faced difficulties with "why"-questions, while they performed better on "when" and "which" questions.

With regard to the OOD-testing LLaMA-3.1-70B Instruct was able to outperform the FT-LMs, fine-tuned on the SQuAD2 dataset, in both F1 and EM-score on three of the five datasets.

Future work could explore techniques to improve prompting specifically for unanswerable questions. It would also be valuable to evaluate the largest model from the LLaMA-3.1 family (405 billion parameters) to determine whether the substantial increase in model size can further close the gap with FT-LM performance.

\bibliography{template}
\bibliographystyle{acm}

\appendix
\section{Additional Results}\label{sec:appendix}
\begin{table}[!ht]
\centering
\caption{\label{tab:results-ftlm}
Full results of the FT-LMs on SQuAD2's official validation dataset based on \href{https://github.com/huggingface/evaluate/blob/main/metrics/squad_v2/compute_score.py}{Huggingface's SQuAD2 Evaluation Skript}
}
\rowcolors{2}{gray!15}{white}

\begin{tabular}{>{\raggedright\arraybackslash}p{2.2cm}>{\centering\arraybackslash}p{2cm}>{\centering\arraybackslash}p{2cm}>{\centering\arraybackslash}p{2cm}}
\rowcolor{gray!30} 
Metric & Flan-T5 & DistilBERT & RoBERTa\\
\hline
exact & 64.549819 & 67.885117 & 79.971364 \\
f1 & 66.019095 & 70.167378 & 82.434992 \\
total & 11873.0 & 11873.0 & 11873.0 \\
HasAns\_exact & 69.416329 & 63.562753 & 76.450742 \\
HasAns\_f1 & 72.359094 & 68.133820 & 81.385064 \\
HasAns\_total & 5928.0 & 5928.0 & 5928.0 \\
NoAns\_exact & 59.697225 & 72.195122 & 83.481918 \\
NoAns\_f1 & 59.697225 & 72.195122 & 83.481918 \\
NoAns\_total & 5945.0 & 5945.0 & 5945.0 \\
best\_exact & 64.659311 & 68.045144 & 80.013476 \\
best\_exact\_thresh & 0.480954 & 0.683446 & 0.841848 \\
best\_f1 & 66.079547 & 70.473248 & 82.470921 \\
best\_f1\_thresh & 0.549976 & 0.749158 & 0.855967 \\
\end{tabular}

\end{table}

\begin{table}[!ht]
\centering
\caption{\label{tab:results-llm-part1}
Results of GPT4-Turbo, LLaMA-2-7B, and LLaMA-3.1-8B Instruct on SQuAD2's official validation dataset}
\rowcolors{2}{gray!15}{white}
\begin{tabular}{>{\raggedright\arraybackslash}p{2.2cm}>{\centering\arraybackslash}p{2.5cm}>{\centering\arraybackslash}p{2.5cm}>{\centering\arraybackslash}p{2.5cm}}
\rowcolor{gray!30} 
Metric & GPT4-Turbo & LLaMA-2-7B & LLaMA-3.1-8B Instruct \\
\hline
exact            & 46.47520 & 43.84739 & 41.00901 \\
f1               & 59.55686 & 50.09121 & 47.22406 \\
total            & 11873 & 11873 & 11873 \\
HasAns\_exact    & 43.82591 & 46.86235 & 68.38731 \\
HasAns\_f1       & 70.02675 & 59.36790 & 80.83524 \\
HasAns\_total    & 5928 & 5928 & 5928. \\
NoAns\_exact     & 49.11691 & 40.84104 & 13.70900 \\
NoAns\_f1        & 49.11691 & 40.84104 & 13.70900 \\
NoAns\_total     & 5945 & 5945 & 5945 \\
best\_exact      & 50.69485 & 50.10528 & 50.11370 \\
best\_f1         & 59.56528 & 51.23518 & 50.30294 \\
\end{tabular}
\end{table}

\begin{table}[!ht]
\centering
\caption{\label{tab:results-llm-part2}
Results of LLaMA-3.1-70B Instruct, LLaMA-3.2 1B Instruct, and LLaMA-3.2 3B Instruct on SQuAD2's official validation dataset}
\rowcolors{2}{gray!15}{white}
\begin{tabular}{>{\raggedright\arraybackslash}p{2.2cm}>{\centering\arraybackslash}p{2.5cm}>{\centering\arraybackslash}p{2.5cm}>{\centering\arraybackslash}p{2.5cm}}
\rowcolor{gray!30} 
Metric & LLaMA-3.1-70B Instruct & LLaMA-3.2 1B Instruct & LLaMA-3.2 3B Instruct \\
\hline
exact            & 57.12962 & 20.97195 & 45.68348 \\
f1               & 73.68071 & 28.27858 & 52.72140 \\
total            & 11873 & 11873 & 11873 \\
HasAns\_exact    & 32.62483 & 40.70513 & 60.07085 \\
HasAns\_f1       & 65.77448 & 55.33934 & 74.16687 \\
HasAns\_total    & 5928 & 5928 & 5928 \\
NoAns\_exact     & 81.56434 & 1.29521 & 31.33726 \\
NoAns\_f1        & 81.56434 & 1.29521 & 31.33726 \\
NoAns\_total     & 5945 & 5945 & 5945 \\
best\_exact      & 57.22227 & 50.09686 & 50.11370 \\
best\_f1         & 73.68201 & 50.09967 & 52.80455 \\
\end{tabular}
\end{table}

\FloatBarrier

\pagebreak
\section{Prompts}

\begin{figure}[!ht]
    \centering
    \includegraphics[scale=0.5]{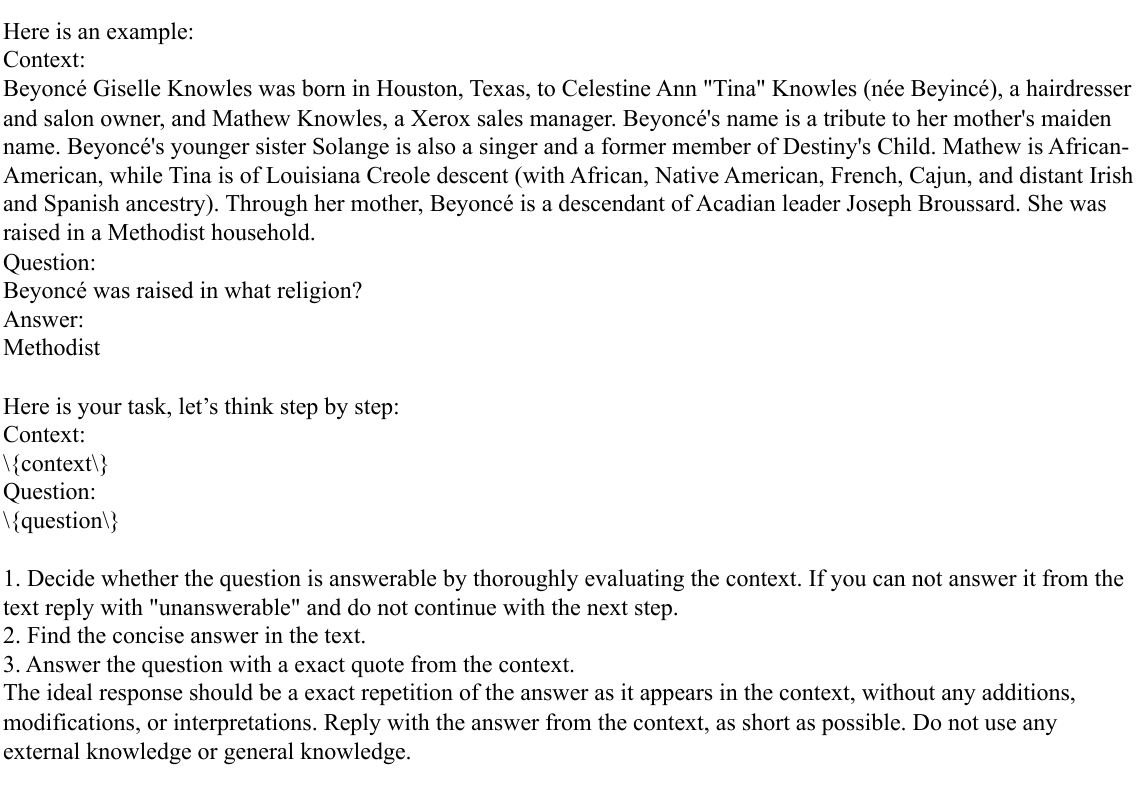}
    \caption{The prompt used for GPT4-Turbo}
    \label{fig:full_prompt_gpt}
\end{figure}

\begin{figure}[!h]
    \centering
    \includegraphics[scale=0.5]{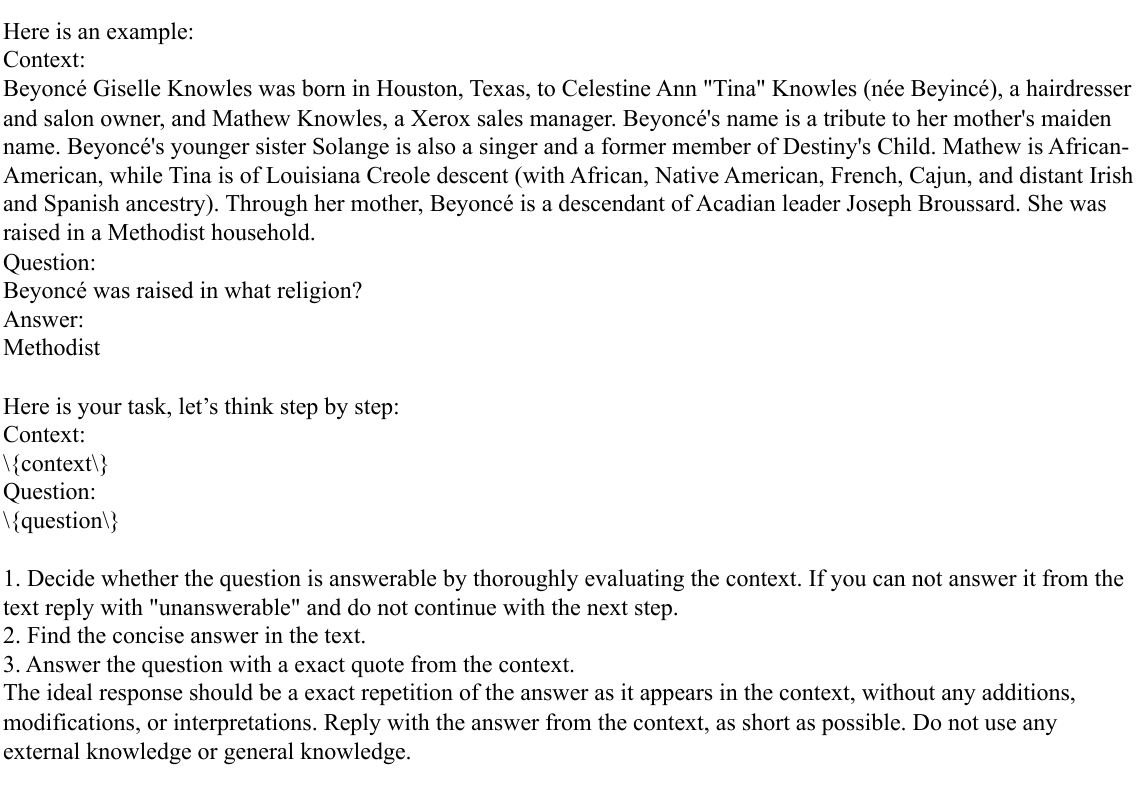}
    \caption{The prompt used for LLaMA-2-7B}
    \label{fig:full_prompt_llama}
\end{figure}

\end{document}